\documentclass[11pt]{article}
\usepackage{colacl}
\title{More accurate tests for the statistical significance of result
differences \thanks{This paper reports on work performed at the MITRE
Corporation under the support of the MITRE Sponsored Research
Program. Warren Greiff, Lynette Hirschman, Christine Doran, John
Henderson, Kenneth Church, Ted Dunning, Wessel Kraaij, Mitch Marcus
and an anonymous reviewer provided helpful suggestions. Copyright
\copyright 2000 The MITRE Corporation. All rights reserved.}  }

\author{Alexander Yeh\\Mitre Corp.\\202 Burlington Rd.\\Bedford, MA 01730\\USA\\asy@mitre.org}

\begin{document}

\maketitle

\typeout{HANDLE COPYRIGHT AND REMOVE PAGE NUMBERS IN FINAL DRAFT!!!}
\typeout{To Save Space, The Bibliography Was Hacked.}

\begin{abstract}
{\begin{picture}(0,0)
\put(0,200){Appears in the 18th International Conference on Computational Linguistics (COLING 2000),}
\put(10,185){pages 947-953, Saarbr\"{u}cken, Germany, July, 2000.}
\put(0,-505){cs.CL/0008005}
\end{picture}}
Statistical significance testing of differences in values of metrics
like recall, precision and balanced F-score is a necessary part of
empirical natural language processing. Unfortunately, we find in a set
of experiments that many commonly used tests often underestimate the
significance and so are less likely to detect differences that exist
between different techniques. This underestimation comes from an
independence assumption that is often violated. We point out some
useful tests that do not make this assumption, including
computationally-intensive randomization tests.
\end{abstract}

\section{Introduction}

In empirical natural language processing, one is often testing whether
some new technique produces improved results (as measured by one or
more metrics) on some test data set when compared to some current
(baseline) technique. When the results are better with the new
technique, a question arises as to whether these result differences
are due to the new technique actually being better or just due to
chance. Unfortunately, one usually cannot directly answer the question
``what is the probability that the new technique is better given the
results on the test data set'':
\begin{center}
P(new technique is better $|$ test set results)
\end{center}
But with statistics, one can answer the following proxy question: if
the new technique was actually no different than the old technique (the
null hypothesis), what is the probability that the results on the test
set would be at least this skewed in the new technique's favor
\cite[Sec.~2.3]{BHH78}? That is, what is
\begin{tabbing}
P(\=test set results at least this skewed\\
  \> in the new technique's favor\\
  \>$|$ new technique is no different than the old)
\end{tabbing}
If the probability is small enough (5\% often is used as the
threshold), then one will reject the null hypothesis and say that the
differences in the results are ``statistically significant'' at that
threshold level.

This paper examines some of the possible methods for trying to detect
statistically significant differences in three commonly used metrics:
recall, precision and balanced F-score. Many of these methods are
found to be problematic in a set of experiments that are
performed. These methods have a tendency to underestimate the
significance of the results, which tends to make one believe that some
new technique is no better than the current technique even when it is.

This underestimate comes from these methods assuming that the
techniques being compared produce independent results when in our
experiments, the techniques being compared tend to produce positively
correlated results.

To handle this problem, we point out some statistical tests, like the
matched-pair $t$, sign and Wilcoxon tests \cite[Sec.~8.7 and
~15.5]{Harnett82}, which do not make this assumption. One can use
these tests on the recall metric, but the precision and balanced
F-score metric have too complex a form for these tests. For such
complex metrics, we use a compute-intensive randomization test
\cite[Sec.~5.3]{Cohen95}, which also avoids this independence
assumption.

The next section describes many of the standard tests used and their
problem of assuming certain forms of independence. The first
subsection describes tests where this assumption appears in estimating
the standard deviation of the difference between the techniques'
results. The second subsection describes using contingency tables and
the $\chi^2$ test. Following this is a section on methods that do not
make this independence assumption. Subsections in turn describe some
analytical tests, how they can apply to recall but not precision or
the F-score, and how to use randomization tests to test precision and
F-score. We conclude with a discussion of dependencies within a test
set's instances, a topic that we have yet to deal with.

\section{Tests that assume independence between compared results}

\subsection{Finding and using the variance of a result difference}\label{s:indep-and-variance}

For each metric, after determining how well a new and current
technique performs on some test set according to that metric, one
takes the difference between those results and asks ``is that
difference significant?''

A way to test this is to expect no difference in the results (the null
hypothesis) and to ask, assuming this expectation, how unusual are
these results? One way to answer this question is to assume that the
difference has a normal or $t$ distribution \cite[Sec.~2.4]{BHH78}. Then
one calculates the following:
\begin{equation}
(d - E[d])/s_d = d/s_d \label{e:t-normal}
\end{equation}
where \mbox{$d=x_1-x_2$} is the difference found between $x_1$ and
$x_2$, the results for the new and current techniques,
respectively. $E[d]$ is the expected difference (which is 0 under the
null hypothesis) and $s_d$ is an estimate of the standard deviation of
$d$. Standard deviation is the square root of the variance, a measure
of how much a random variable is expected to vary. The results of
equation~\ref{e:t-normal} are compared to tables (c.f. in
\newcite[Appendix]{BHH78}) to find out what the chances are of
equaling or exceeding the equation~\ref{e:t-normal} results if the
null hypothesis were true. The larger the equation~\ref{e:t-normal}
results, the more unusual it would be under the null hypothesis.

A complication of using equation~\ref{e:t-normal} is that one usually
does not have $s_d$, but only $s_1$ and $s_2$, where $s_1$ is the
estimate for $x_1$'s standard deviation and similarly for $s_2$. How
does one get the former from the latter? It turns out that
\cite[Ch.~3]{BHH78}
\begin{displaymath}
\sigma^2_d = \sigma^2_1 + \sigma^2_2 - 2\rho_{12}\sigma_1\sigma_2
\end{displaymath}
where $\sigma_i$ is the true standard deviation (instead of the
estimate $s_i$) and $\rho_{12}$ is the correlation coefficient between
$x_1$ and $x_2$. Analogously, it turns out that 
\begin{equation}
s^2_d = s^2_1 + s^2_2 - 2r_{12}s_1s_2 \label{e:var-dif}
\end{equation}
where $r_{12}$ is an estimate for $\rho_{12}$. So not only does
$\sigma_d$ (and $s_d$) depend on the properties of $x_1$ and $x_2$ in
isolation, it also depends on how $x_1$ and $x_2$ interact, as
measured by $\rho_{12}$ (and $r_{12}$). When $x_1$ and $x_2$ are
independent, $\rho_{12}=0$, and then \mbox{$\sigma_d =
\sqrt{\sigma^2_1 + \sigma^2_2}$} and analogously, \mbox{$s_d =
\sqrt{s^2_1 + s^2_2}$}. When $\rho_{12}$ is positive, $x_1$ and $x_2$
are positively correlated: a rise in $x_1$ or $x_2$ tends to be
accompanied by a rise in the other result. When $\rho_{12}$ is
negative, $x_1$ and $x_2$ are negatively correlated: a rise in $x_1$
or $x_2$ tends to be accompanied by a decline in the other
result. \mbox{$-1\leq\rho_{12}\leq 1$} \cite[Sec.~10.2]{LandM86}.

%xbar = sum[x]/N, Vhat(x) = s^2 for x
%Vhat(x-y) = sum[((x-y)-(x-y)bar)^2]/(N-1) = sum[(x-xbar-y+ybar)^2]/(N-1)
% = sum[x^2-2x(xbar)-2xy+2x(ybar)+(xbar)^2+2(xbar)y-2(xbar)ybar+y^2-2y(ybar)+(ybar)^2]/(N-1)
% = (sum[x^2]-2N(xbar)^2-2sum[xy]+2N(xbar)ybar+N(xbar)^2+2N(xbar)ybar-2N(xbar)ybar+sum(y^2)-2N(ybar)^2+N(ybar)^2)/2
% = (sum[x^2]-2N(xbar)^2-2sum[xy]+2N(xbar)ybar+N(xbar)^2+0+sum(y^2)-N(ybar)^2)/(N-1)
% = (sum[x^2]-2N(xbar)^2+N(xbar)^2+sum(y^2)-N(ybar)^2-2sum[xy]+2N(xbar)ybar)/(N-1)
% = ((sum[x^2]-N(xbar)^2)+(sum(y^2)-N(ybar)^2)-2(sum[xy]-N(xbar)ybar))/(N-1)
% = Vhat(x)+Vhat(y)-2COVhat(xy)
% r = COVhat(xy)/sqrt[Vhat(x)Vhat(y)]

The assumption of independence is often used in formulas to determine
the statistical significance of the difference $d=x_1-x_2$. But how
accurate is this assumption? One might expect some positive
correlation from both results coming from the same test set. One may
also expect some positive correlation when either both techniques are
just variations of each other\footnote{These variations are often
designed to usually behave in the same way and only differ in just a
few cases.} or both techniques are trained on the same set of training
data (and so are missing the same examples relative to the test set).

This assumption was tested during some experiments for finding
grammatical relations (subject, object, various types of modifiers,
etc.). The metric used was the fraction of the relations of interest
in the test set that were {\em recall}\/ed (found) by some
technique. The relations of interest were various subsets of the 748
relation instances in that test set. An example subset is all the
modifier relations. Another subset is just that of all the time
modifier relations.

First, two different techniques, one memory-based and the other
transformation-rule based, were trained on the same training set, and
then both tested on that test set. Recall comparisons were made for
ten subsets of the relations and the $r_{12}$ was found for each
comparison. From \newcite[Ch.~3]{BHH78}
\begin{displaymath}
r_{12} = \sum_k(y_{1k}-\overline{y}_1)(y_{2k}-\overline{y}_2)/(s_1s_2(n-1))
\end{displaymath}
where $y_{ik}=1$ if the $i$\/th technique recalls the $k$\/th relation
and $=0$ if not. $n$ is the number of relations in the subset.
$\overline{y}_i$ and $s_i$ are mean and standard deviation estimates
(based on the $y_{ik}$'s), respectively, for the $i$\/th technique.

For the ten subsets, only one comparison had a $r_{12}$ close to 0 (It
was \mbox{-0.05}). The other nine comparisons had $r_{12}$'s between
0.29 and 0.53. The ten comparison median value was 0.38.

Next, the transformation-rule based technique was run with different
sets of starting conditions and/or different, but overlapping, subsets
of the training set. Recall comparisons were made on the same test
data set between the different variations. Many of the comparisons
were of how well two variations recalled a particular subset of the
relations. A total of 40 comparisons were made. The $r_{12}$'s on all
40 were positive. 3 of the $r_{12}$'s were in the 0.20-0.30 range. 24
of the $r_{12}$'s were in the 0.50-0.79 range. 13 of the $r_{12}$'s
were in the 0.80-1.00 range.

So in our experiments, we were usually comparing positively
correlated results. How much error is introduced by assuming
independence? An easy-to-analyze case is when the standard deviations
for the results being compared are the same.\footnote{This is actually
roughly true in the comparisons made, and is assumed to be true in
many of the standard tests for statistical significance.} Then
equation~\ref{e:var-dif} reduces to \mbox{$s_d = s \sqrt{2(1 -
r_{12})}$}, where $s = s_1 = s_2$. If one
assumes the results are independent (assume \mbox{$r_{12}=0$}), then
\mbox{$s_d = s \sqrt{2}$}. Call this value $s_{d-ind}$. As $r_{12}$
increases in value, $s_d$ decreases:
\begin{center}$
\begin{array}{|l|r|r|}\hline
r_{12} & s_d & (s_{d-ind})/s_d \\ \hline
0.38 & 0.787(s_{d-ind})& 1.27 \\ \hline
0.50 & 0.707(s_{d-ind})& 1.41 \\ \hline
0.80 & 0.447(s_{d-ind})& 2.24 \\ \hline
\end{array}$
\end{center}
The rightmost column above indicates the magnitude by which
erroneously assuming independence (using $s_{d-ind}$ in place of
$s_d$) will increase the standard deviation estimate. In
equation~\ref{e:t-normal}, $s_d$ forms the denominator of the ratio
$d/s_d$. So erroneously assuming independence will mean that the
numerator $d$, the difference between the two results, will need to
increase by that same factor in order for equation~\ref{e:t-normal} to
have the same value as without the independence assumption. Since the
value of that equation indicates the statistical significance of $d$,
assuming independence will mean that $d$ will have to be larger than
without the assumption to achieve the same apparent level of
statistical significance. From the table above, when
\mbox{$r_{12}=0.50$}, $d$ will need to be about 41\% larger. Another
way to look at this is that assuming independence will make the same
value of $d$ appear less statistically significant.

The common tests of statistical significance use this assumption. The
test known as the $t$ \cite[Sec.~4.1]{BHH78} or two-sample $t$
\cite[Sec.~8.7]{Harnett82} test does. This test uses
equation~\ref{e:t-normal} and then compares the resulting value
against the $t$ distribution tables. This test has a
complicated form for $s_d$ because:
\begin{enumerate}
 \item $x_1$ and $x_2$ can be based on differing numbers of
 samples. Call these numbers $n_1$ and $n_2$ respectively.

 \item In this test, the $x_i$'s are each an $n_i$ sample average of
 another variable (call it $y_i$). This is important because the
 $s_i$'s in this test are standard deviation estimates for the
 $y_i$'s, not the $x_i$'s. The relationship between them is that $s_i$
 for $y_i$ is the same as \mbox{$(\sqrt{n_i})s_i$} for $x_i$.

 \item The test itself assumes that $y_1$ and $y_2$ have the same
 standard deviation
 (call this common value $s$). The denominator estimates
 $s$ using a weighted average of $s_1$ and $s_2$. The weighting is
 based on $n_1$ and $n_2$.
\end{enumerate}
From \newcite[Sec.~8.7]{Harnett82}, the denominator
\begin{displaymath}
s_d = \sqrt{\left(\frac{(n_1-1)s_1^2+(n_2-1)s_2^2}{n_1+n_2-2}\right)\left(\frac{n_1+n_2}{n_1n_2}\right)}
\end{displaymath}
When \mbox{$n_1=n_2$} (call this common value $n$), $s_1$ and $s_2$
will be given equal weight, and $s_d$ simplifies to
\mbox{$\sqrt{(s_1^2+s_2^2)/n}$}. Making the substitution described
above of \mbox{$s_i\sqrt{n_i}$} for $s_i$ leads to an $s_d$ of
\mbox{$\sqrt{s_1^2+s_2^2}$}, the form we had earlier for using the
independence assumption.

Another test that both makes this assumption and uses a form of
equation~\ref{e:t-normal} is a test for binomial data
\cite[Sec.~8.11]{Harnett82} which uses the ``fact'' that binomial
distributions tend to approximate normal distributions. In this test,
the $x_i$'s being compared are the fraction of the items of interest
that are recovered by the $i$th technique. In this test, the
denominator $s_d$ of equation~\ref{e:t-normal} also has a complicated
form, both due to the reasons mentioned for the $t$ test above and to
the fact that with a binomial distribution, the standard deviation is
a function of the number of samples and the mean value.

\subsection{Using contingency tables and $\chi^2$ to test precision}\label{ss:chi-square}

A test that does {\em not} use equation~\ref{e:t-normal} but still
makes an assumption of independence between $x_1$ and $x_2$ is that of
using contingency tables with the chi-squared ($\chi^2$) distribution
\cite[Sec.~5.7]{BHH78}. When the assumption is valid, this test is
good for comparing differences in the {\em precision}
metric. Precision is the fraction of the items ``found'' by some
technique that are actually of interest. Precision \mbox{$= R/(R+S)$},
where $R$ is the number of items that are of interest and are Recalled
(found) by the technique, and $S$ is the number of items that are
found by the technique that turn out to be Spurious (not of
interest). One can test whether the precision results from two techniques
are different by using a $2\times 2$ contingency table to test whether the
ratio $R/S$ is different for the two techniques. One makes the latter
test by seeing if the assumption that the ratios for the two
techniques are the same (the null hypothesis) leads to a statistically
significant result when using a $\chi^2$ distribution with one degree
of freedom. A $2\times 2$ table has 4 cells. The top 2 cells are
filled with the $R$ and $S$ of one technique and the bottom 2 cells
get the $R$ and $S$ of the other technique. In this test, the value in
each cell is assumed to have a Poisson distribution. When the cell
values are not too small, these Poisson distributions are
approximately Normal (Gaussian). As a result, when the cell values are
independent, summing the normalized squares of the difference between
each cell and its expected value leads to a $\chi^2$ distribution
\cite[Sec.~2.5-2.6]{BHH78}.

How well does this test work in our experiments? Precision is a
non-linear function of two random variables $R$ and $S$, so we did not
try to estimate the correlation coefficient for precision. However, we
can easily estimate the correlation coefficients for the $R$'s. They
are the $r_{12}$'s found in section~\ref{s:indep-and-variance}. As
that section mentions, the $r_{12}$'s found are just about always
positive. So at least in our experiments, the $R$'s are not
independent, but are positively correlated, which violates the
assumptions of the test.

An example of how this test behaves is the following comparison of the
precision of two different methods at finding the modifier relations
using the same training and test set. The correlation coefficient
estimate for $R$ is 0.35 and the data is
\begin{center}
\begin{tabular}{|l|r|r|r|}\hline
Method & $R$ & $S$ & Precision\\ \hline
 1 & 47 & 48 & 49\%\\ \hline
 2 & 25 & 14 & 64\% \\ \hline
\end{tabular}
\end{center}
%(/ (float (* (- (* 47. 14.) (* 48. 25.)) (- (* 47. 14.) (* 48. 25.)) (+ 47. 48. 25. 14.))) (+ 47. 48.) (+ 25. 14.) (+ 47. 25.) (+ 48. 14.))
Placing the $R$ and $S$ values into a $2\times 2$ table leads to a
$\chi^2$ value of 2.38.\footnote{We do {\em not} use Yate's
adjustment to compensate for the numbers in the table being
integers. Doing so would have made the results even worse.}
 At 1 degree of freedom, the $\chi^2$ tables
indicate that if the null hypothesis were true, there would be a 10\% to
20\% chance of producing a $\chi^2$ value at least this large. So
according to this test, this much of an observed difference in
precision would not be unusual if no actual difference in the
precision exists between the two methods. 

This test assumes independence between the $R$ values.  When we use a
$2^{20}$ (=1048576) trial approximate randomization test
(section~\ref{ss:randomization}), which makes no such assumptions,
then we find that this latter test indicates that under the null
hypothesis, there is less than a 4\% chance of producing a difference
in precision results as large as the one observed. So this latter test
indicates that this much of an observed difference in precision would
be unusual if no actual difference in the precision exists between the
two methods.

It should be mentioned that the manner of testing here is slightly
different than the manner in the rest of this paper.  The $\chi^2$
test looks at the square of the difference of two results, and rejects
the null hypothesis (the compared techniques are the same) when this
square is large, whether the largeness is caused by the new technique
producing a much better result than the current technique or
vice-versa. So to be fair, we compared the $\chi^2$ results with a
two-sided version of the randomization test: estimate the likelihood
that the observed magnitude of the result difference would be matched
or exceeded (regardless of which technique produced the better result)
under the null hypothesis. A one-sided version of the test, which is
comparable to what we use in the rest of the paper, estimates the
likelihood of a different outcome under the null hypothesis: that of
matching or exceeding the difference of how much better the new
(possibly better) technique's observed result is than the current
technique's observed result. In the above scenario, a one-sided test
produces a 3\% figure instead of a 4\% figure.

\section{Tests without that independence assumption}

\subsection{Tests for matched pairs}\label{ss:matched-pair-tests}

At this point, one may wonder if all statistical tests make such an
independence assumption. The answer is no, but those tests that do not
measure how much two techniques interact do need to make some
assumption about that interaction and typically, that assumption is
independence. Those tests that notice in some way how much two
techniques interact can use those observations instead of relying on
assumptions.

One way to measure how two techniques interact is to compare how
similarly the two techniques react to various parts of the test
set. This is done in the matched-pair $t$ test
\cite[Sec.~8.7]{Harnett82}. This test finds the difference between how
techniques~1 and~2 perform on each test set sample. The $t$
distribution and a form of equation~\ref{e:t-normal} are used. The
null hypothesis is still that the numerator $d$ has a 0 mean, but $d$
is now the sum of these difference values (divided by the number of
samples), instead of being \mbox{$x_1-x_2$}. Similarly, the
denominator $s_d$ is now estimating the standard deviation of these
difference values, instead of being a function of $s_1$ and
$s_2$. This means for example, that even if the values from
techniques~1 and~2 vary on different test samples, $s_d$ will now be 0
if on each test sample, technique~1 produces a value that is the same
constant amount more than the value from technique~2.

Two other tests for comparing how two techniques perform by comparing
how well they perform on each test sample are the sign and Wilcoxon
tests \cite[Sec.~15.5]{Harnett82}. Unlike the matched-pair $t$ test,
neither of these two tests assume that the sum of the differences has a
normal (Gaussian) distribution. The two tests are so-called
nonparametric tests, which do not make assumptions about how the
results are distributed \cite[Ch.~15]{Harnett82}.

The sign test is the simplier of the two. It uses a binomial
distribution to examine the number of test samples where technique~1
performs better than technique~2 versus the number where the opposite
occurs. The null hypothesis is that the two techniques perform equally
well.

Unlike the sign test, the Wilcoxon test also uses information on how
large a difference exists between the two techniques' results on each
of the test samples.

\subsection{Using the tests for matched-pairs}

All three of the matched-pair $t$, sign and Wilcoxon tests can be
applied to the recall metric, which is the fraction of the items of
interest in the test set that a technique recalls (finds). Each item
of interest in the test data serves as a test sample. We use the sign
test because it makes fewer assumptions than the matched-pair $t$ test
and is simplier than the Wilcoxon test. In addition, the fact that the
sign test ignores the size of the result difference on each test
sample does not matter here. With the recall metric, each sample of
interest is either found or not by a technique. There are no
intermediate values.

While the three tests described in section~\ref{ss:matched-pair-tests}
can be used on the recall metric, they cannot be straightforwardly
used on either the precision or balanced F-score metrics. This is
because both precision and F-score are more complicated non-linear
functions of random variables than recall. In fact both can be thought
of as non-linear functions involving recall. As described in
Section~\ref{ss:chi-square}, precision \mbox{$= R/(R+S)$}, where $R$
is the number of items that are of interest that are {\em recall}\/ed
by a technique and $S$ is the number of items (found by a technique)
that are not of interest. The balanced F-score \mbox{$=2ab/(a+b)$},
where $a$ is recall and $b$ is precision.

\subsection{Using randomization for precision and F-score}\label{ss:randomization}

A class of technique that can handle all kinds of functions of random
variables without the above problems is the computationally-intensive
randomization tests \cite[Ch.~2]{Noreen89} \cite[Sec.~5.3]{Cohen95}.
These tests have previously used on such functions during the
``message understanding'' (MUC) evaluations \cite{CHL93}.  The
randomization test we use is like a randomization version of the
paired sample (matched-pair) $t$ test \cite[Sec.~5.3.2]{Cohen95}. This
is a type of stratified shuffling \cite[Sec.~2.7]{Noreen89}. When
comparing two techniques, we gather-up all the responses (whether
actually of interest or not) produced by one of the two techniques
when examining the test data, but not both techniques. Under the null
hypothesis, the two techniques are not really different, so any
response produced by one of the techniques could have just as likely
come from the other. So we shuffle these responses, reassign each
response to one of the two techniques (equally likely to either
technique) and see how likely such a shuffle produces a difference
(new technique minus old technique) in the metric(s) of interest (in
our case, precision and F-score) that is at least as large as the
difference observed when using the two techniques on the test data.

$n$ responses to shuffle and assign\footnote{Note that responses
produced by both or neither techniques do not need to be shuffled and
assigned.} leads to $2^n$ different ways to shuffle and assign those
responses. So when $n$ is small, one can try each of the different
shuffles once and produce an exact randomization. When $n$ gets large,
the number of different shuffles gets too large to be exhaustively
evaluated. Then one performs an approximate randomization where each
shuffle is performed with random assignments.

For us, when \mbox{$n\leq 20$} (\mbox{$2^n \leq 1048576$}), we use an
exact randomization. For \mbox{$n> 20$}, we use an approximate
randomization with 1048576 shuffles. Because an approximate
randomization uses random numbers, which both lead to occasional
unusual results and may involve using a not-so-good pseudo-random
number generator\footnote{One example is the RANDU routine on the
IBM360 \cite[Sec.~10.1]{FMM77}.}, we perform the following checks:
\begin{itemize}
\item We run the 1048576 shuffles a second time and compare the two
sets of results.

\item We also use the same shuffles to calculate the statistical
significance for the recall metric, and compare this significance
value with the significance value found for recall analytically by the
sign test.
\end{itemize}

An example of using randomization is to compare two different methods
on finding modifier relations in the same test set.
The results on the test set are:
\begin{center}
\begin{tabular}{|l|r|r|r|}\hline
Method & Recall & Precision & F-score\\ \hline
  I   &    45.6\% & 49.5\% &  47.5\%\\ \hline
  II   &    24.3\% & 64.1\% &  35.2\%\\ \hline
\end{tabular}
\end{center}
Two questions being tested are whether the apparent improvement in
recall and F-score from using method~I is significant. Also being
tested is whether the apparent improvement in precision from using
method~II is significant.

In this example, there are 103 relations that should be found (are of
interest). Of these, 19 are recalled by both methods, 28 are
recalled by method I but not II, and 6 are recalled by II but
not I. The correlation coefficient estimate between the methods'
recalls is 0.35.  In addition, 5 spurious (not of interest) relations
are found by both methods, with method I finding an additional
43 spurious relationships (not found by method II) and method II
finding an additional~9 relationships.

There are a total of 28+6+43+9=86 relations that are found (whether of
interest or not) by one method, but not the other. This is too many
to perform an exact randomization, so a 1048576 trial approximate
randomization is performed. 

In 96 of these trials, method~I's recall is greater than method~II's
recall by at least (45.6\%$-$24.3\%). Similarly, in 14794 of the
trials, the F-score difference is at least (47.5\%$-$35.2\%).  In
25770 of the trials, method~II's precision is greater than method~I's
precision by at least (64.1\%$-$49.5\%). From
\cite[Sec.~3A.3]{Noreen89}, the significance level (probability under
the null hypothesis) is at most \mbox{$(nc+1)/(nt+1)$}, where $nc$ is
the number of trials that meet the criterion and $nt$ is the number of
trials. So for recall, the significance level is at most
\mbox{(96+1)/(1048576+1)} =0.00009. Similarly, for F-score, the
significance level is at most 0.014 and for precision, the level is at
most 0.025. A second 1048576 trial produces similar results, as does a
sign test on recall. Thus, we see that all three differences are
statistically significant.

\section{The future: handling inter-sample dependencies}

An assumption made by all the methods mentioned in this paper is that
the members of the test set are all independent of one another. That
is, knowing how a method performs on one test set sample should not
give any information on how that method performs on other test set
samples. This assumption is not always true.

\newcite{CandM93} give some examples of dependence between test set
instances in natural language. One type of dependence is that of a
lexeme's part of speech on the parts of speech of neighboring lexemes
(their section~2.1). Similar is the concept of {\em collocation},
where the probability of a lexeme's appearance is influenced by the
lexemes appearing in nearby positions (their section~3). A type of
dependence that is less local is that often, a content word's
appearance in a piece of text greatly increases the chances of that
same word appearing later in that piece of text (their section~2.3).

What are the effects when some dependency exists? The expected
(average) value of the instance results will stay the same. However,
the chances of getting an unusual result can change. As an example,
take five flips of a fair coin. When no dependencies exist between the
flips, the chances of the extreme result that all the flips land on a
particular side is fairly small ($(1/2)^5=1/32$). When the flips are
positively correlated, these chances increase. When the first flip
lands on that side, the chances of the other four flips doing
the same are now each greater than $1/2$.

Since statistical significance testing involves finding the chances of
getting an unusual (skewed) result under some null hypothesis, one
needs to determine those dependencies in order to accurately determine
those chances. Determining the effect of these dependencies is
something that is yet to be done.

\section{Conclusions}

In empirical natural language processing, one is often comparing
differences in values of metrics like recall, precision and balanced
F-score. Many of the statistics tests commonly used to make such
comparisons assume the independence between the results being
compared. We ran a set of natural language processing experiments and
found that this assumption is often violated in such a way as to
understate the statistical significance of the differences between the
results. We point out some analytical statistics tests like
matched-pair $t$, sign and Wilcoxon tests, which do not make this
assumption and show that they can be used on a metric like recall. For
more complicated metrics like precision and balanced F-score, we use a
compute-intensive randomization test, which also avoids this
assumption. A next topic to address is that of possible dependencies
between test set samples.

%\bibliographystyle{acl}

%\bibliography{nl}

\begin{thebibliography}{}

\bibitem[\protect\citename{Box \bgroup et al.\egroup }1978]{BHH78}
G.~Box, W.~Hunter, and J.~Hunter.
\newblock 1978.
\newblock {\em Statistics for experimenters}.
\newblock John Wiley and Sons.

\bibitem[\protect\citename{Chinchor \bgroup et al.\egroup }1993]{CHL93}
N.~Chinchor, L.~Hirschman, and D.~Lewis.
\newblock 1993.
\newblock Evaluating message understanding systems: an analysis of the third
  message understanding conference (muc-3).
\newblock {\em Computational Linguistics}, 19(3).

\bibitem[\protect\citename{Church and Mercer}1993]{CandM93}
K.~Church and R.~Mercer.
\newblock 1993.
\newblock Introduction to the special issue on computational linguistics using
  large corpora.
\newblock {\em Computational Linguistics}, 19(1):1--24.

\bibitem[\protect\citename{Cohen}1995]{Cohen95}
P.~Cohen.
\newblock 1995.
\newblock {\em Empirical Methods for Artificial Intelligence}.
\newblock MIT Press, MA, USA.

\bibitem[\protect\citename{Forsythe \bgroup et al.\egroup }1977]{FMM77}
G.~Forsythe, M.~Malcolm, and C.~Moler.
\newblock 1977.
\newblock {\em Computer methods for mathematical computations}.
\newblock Prentice-Hall, NJ, USA.

\bibitem[\protect\citename{Harnett}1982]{Harnett82}
D.~Harnett.
\newblock 1982.
\newblock {\em Statistical Methods}.
\newblock Addison-Wesley Publishing Co., 3rd edition.

\bibitem[\protect\citename{Larsen and Marx}1986]{LandM86}
R.~Larsen and M.~Marx.
\newblock 1986.
\newblock {\em An Introduction to Mathematical Statistics and Its
  Applications}.
\newblock Prentice-Hall, NJ, USA, 2nd edition.

\bibitem[\protect\citename{Noreen}1989]{Noreen89}
E.~Noreen.
\newblock 1989.
\newblock {\em Computer-intensive methods for testing hypotheses: an
  introduction}.
\newblock John Wiley and Sons, Inc.

\end{thebibliography}

\end{document}